\documentclass[10pt, a4paper]{article}
\usepackage{lrec-coling2024} 

\usepackage{enumerate}
\usepackage{graphicx, float}
\usepackage{amsmath}
\usepackage{amsfonts}
\usepackage{amssymb}
\usepackage{multirow}
\usepackage{tablefootnote}
\usepackage{subcaption}
\usepackage{comment}
\hypersetup{pdfauthor={Meliksah Turker, Mehmet Erdi Ari, Aydin Han},
            pdftitle={VBART: The Turkish LLM},
            }

\title{VBART: The Turkish LLM}

\name{Meliksah Turker, Mehmet Erdi Ari, Aydin Han} 
\address{VNGRS-AI \\
         YTÜ Teknopark B2 103 Davutpaşa, İstanbul, Turkey \\
         \{meliksah.turker, erdi.ari, aydin.han\}@vngrs.com\\}

\abstract{
We present VBART, the first Turkish sequence-to-sequence Large Language Models (LLMs) pre-trained on a large corpus from scratch.
VBART are compact LLMs based on good ideas leveraged from BART and mBART models and come in two sizes, Large and XLarge.
Fine-tuned VBART models surpass the prior state-of-the-art results in abstractive text summarization, title generation, text paraphrasing, question answering and question generation tasks.
They allow fine-tuning for future text generation tasks and datasets, carving a new path for Turkish Natural Language Processing (NLP) research.
Our work shows that having a pre-trained LLM for Turkish outperforms up to 3x multilingual models, improving existing results and providing efficient models for training and inference.
Moreover, we show that our monolingual tokenizer is up to 11x more efficient than multilingual tokenizers.
Last but not least, we introduce a method to enlarge an existing pre-trained LLM and question the relevancy of Chinchilla Scaling Law to sequence-to-sequence masked language models.
Our fine-tuned models, tokenizer and cleaned vngrs-web-corpus of 135 GB are publicly available at \href{https://huggingface.co/vngrs-ai}{huggingface.co/vngrs-ai}.
\\ \newline 
\Keywords{
Large Language Models, LLM, Turkish, BART, sequence-to-sequence, Text Summarization, Title Generation, Text Paraphrasing, Question Generation, Question Answering.
}
}

\begin{document}

\maketitleabstract

\section{Introduction} 
\label{sec:intro}
Natural Language Processing (NLP) research landscape has changed drastically over the last decade.
Invention of word embedding methods like Word2Vec~\cite{word2vec_mikolov2013efficient} and GloVe~\cite{glove_pennington2014glove} has been the foundation of transfer learning.
They have been succeeded by FastText~\cite{fasttext_bojanowski2017enriching}, which leverages character n-grams, and ELMo~\cite{elmo_peters2018deep}, which uses the context of the words for the first time.

Using pre-trained word embeddings for text classification tasks often involved Recurrent Neural Networks (RNNs) such as Long Short-Term Memory (LSTM)~\cite{lstm_hochreiter1997long} and Gated Recurrent Unit (GRU)~\cite{gru_cho2014learning}.
Democratization of these models has been possible thanks to Deep Learning frameworks like Keras~\cite{keras_chollet2015keras}, Tensorflow~\cite{tensorflow_tensorflow2015} and PyTorch~\cite{pytorch_NEURIPS2019_9015}.

Tokenization methods such as WordPiece~\cite{wordpiece_wu2016google}, Byte-Pair Encoding (BPE)~\cite{bpe_sennrich2015neural} and Unigram model~\cite{unigram_kudo2018subword} solved the out-of-vocabulary problem while keeping the vocabulary size feasible.

Transformer~\cite{attentionisallyouneed_vaswani2017attention} architecture and the introduction of BERT~\cite{bert_devlin2018bert} have made a significant contribution to the field of transfer learning.
BERT allowed for achieving state-of-the-art results efficiently.

Following the success of BERT, an Encoder-only model, BART~\cite{bart_lewis2019bart} and T5~\cite{t5_2019t5} models have shown that it is possible to apply the principle of unsupervised pre-training of Masked Language Models to seq2seq tasks for conditional text generation.

They are followed by multilingual versions mBART~\cite{mbart_liu2020multilingual} and mT5~\cite{mt5mc4_xue2020mt5}, pre-training of which are conducted on multiple languages together.
This allowed them to be fine-tuned for low-resource languages without suffering the high cost of pre-training a new model for the language from scratch.
State-of-the-art results in conditional text generation tasks are obtained~\cite{baykara_seq2seq_baykara2022turkish, mukayese_safaya2022mukayese, qaqg_akyon2022questgen} for Turkish by fine-tuning them.

\begin{figure*}[]
\centering
\includegraphics[width=1.3\columnwidth]{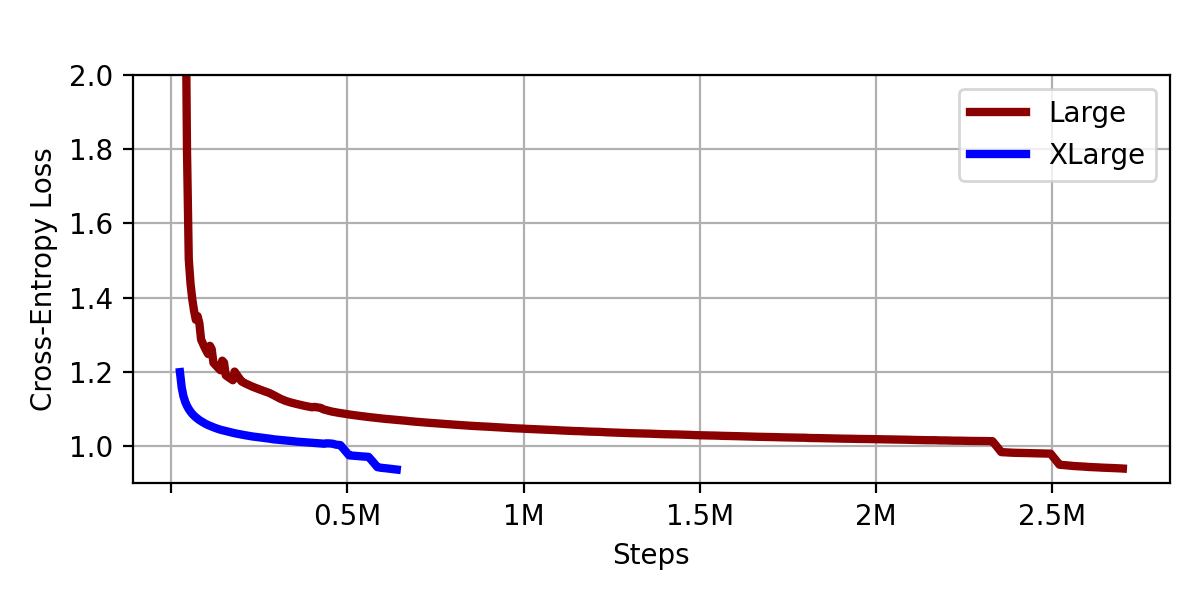}
\caption{Pre-training loss of VBART models.
Note that there are two steep drops in loss towards the end of the training.
This is due to the reduction in Dropout from 0.10 to 0.05 and then to 0.}
\label{fig:train_loss}
\end{figure*}

Lastly, very large language models like BLOOM~\cite{bloom_scao2022bloom}, PaLM~\cite{palm_chowdhery2022palm}, GPT-4~\cite{gpt4_tr_OpenAI2023GPT4TR} have made a significant impact on both the industry and the academy.
They achieved state-of-the-art results in English and high-resource languages and decent results for low-resource languages.
However, as in the case of ChatGPT~\cite{chatgpt}, distilling very large models results in poor performance~\cite{chatgpt_lrl_lai2023chatgpt, chatgpt_translator_jiao2023chatgpt} for low-resource languages.
On the other hand, using very large models is computationally expensive.

It is clear that the performance of multilingual LLMs in low-resource languages is a matter of trade-off between the cost of computation and evaluation metrics.
This is because they are not optimal for a specific language since a significant portion of training time and network capacity are spent on other languages.
This is especially true for low-resource languages such as Turkish.
Hence, a dedicated LLM is needed to obtain state-of-the-art results in an efficient and computationally cheap manner for a given language.

Thus, we present VBART, the first dedicated seq2seq LLM pre-trained for Turkish on a large scale.
First, we train VBART-Large from scratch on a cleaned corpus of 135.7 GBs of Turkish text based on the pre-training task of BART and the model architecture of mBART.
Then, we create VBART-XLarge by enlarging the prior model by doubling the number of encoder and decoder layers, using the weights of VBART-Large.
Both models surpass the previous state-of-the-art in abstractive text summarization, paraphrasing, title generation, question answering and question generation tasks.
Our contributions to Turkish NLP research are as follows:

\begin{enumerate}
    \item SentencePiece Unigram Tokenizer
    \item vngrs-web-corpus that consist of 135 GBs of cleaned text
    \item VBART LLM Models
    \item New state-of-the-art results for text generation tasks in Turkish.
\end{enumerate}

\section{Related Work}
The existence of BERTurk~\cite{berturk}, a dedicated pre-trained Encoder only Language Model (LM) for Turkish allowed obtaining state-of-the-art results on text~\cite{bert_tr_small_toprak2023developing, berturk_offensive_ozdemir2020nlp, berturk_sentiment_koksal2021twitter, berturk_topicdetection_csahinucc2021topic} and token~\cite{berturk_ner_aras2021evaluation, berturk_ner_ccarik2022twitter} classification tasks.

In the case of text generation tasks, however, researchers are compelled to either use BERTurk-based hybrid solutions or fine-tune pre-trained multilingual seq2seq LLM like mBART and mT5 since there is no dedicated seq2seq LLM for Turkish.

\begin{table*}[]
\centering
\begin{tabular}{l c c c c}
	\scriptsize R1/R2/RL & Parameters & MLSum & TRNews & XLSum \\
	\hline
        \multicolumn{5}{l}{\textbf{~\citealp{mukayese_safaya2022mukayese}}} \\%
        \hline
        Base &120M & 40.23/27.23/35.08 & & \\
        mBART50 & 610M & 43.75/30.60/38.47 & & \\
        mT5-Base & 580M & 44.13/31.09/38.76 & & \\
        \hline
        \multicolumn{5}{l}{\textbf{~\citealp{baykara_seq2seq_baykara2022turkish}}} \\
        \hline
        BERT2BERT-32K & 248M & 41.48/27.23/37.66 & 41.06/25.60/\textbf{37.69} & \\
        mBART25 & 610M & 40.47/26.17/36.22 & 40.52/25.22/36.80 & \\
        mT5-Base & 580M & 42.26/27.81/37.96 & 41.13/25.75/37.60 & \\
        \hline
        \multicolumn{5}{l}{\textbf{Our work}} \\
        \hline
        VBART-Large & 387M & \textbf{45.75}/\textbf{32.71}/\textbf{39.86} & \textbf{41.97}/\textbf{28.26}/36.69 & 34.15/17.94/28.03\\
        VBART-XLarge & 740M & \textbf{46.13}/\textbf{33.01}/\textbf{40.42} & \textbf{42.55}/\textbf{28.69}/37.42 & 35.13/18.80/29.18\\
        \hline

\end{tabular}
\caption{Text Summarization}
\label{tbl:sum}
\end{table*}

\textbf{Text Summarization:}
\citet{baykara_pointer_baykara2022abstractive} has built a seq2seq model by combining a pre-trained BERT Encoder and a randomly initialized Transformer Decoder.
\citet{baykara_seq2seq_baykara2022turkish} and
\citet{mukayese_safaya2022mukayese} have fine-tuned mBART and mT5 on text summarization datasets.

\textbf{Title Generation}
While working on the internet news domain for text summarization, \citet{baykara_pointer_baykara2022abstractive, baykara_seq2seq_baykara2022turkish} have created a dataset of news titles and fine-tuned mBART and mT5 models to generate title from the given the news summary.

\textbf{Question Generation \& Answering:}
\citet{qaqg_akyon2022questgen} has fine-tuned mT5 for question generation and answering tasks.

\textbf{Text Paraphrasing:}
\citet{bert2bert_paraphrase} used pre-trained BERT weights to initialize both encoder and decoder weights to create a seq2seq transformer network for text paraphrasing task.

Even though using BERT weights to initialize encoder weights is a reasonable approach, initializing the decoder network weights randomly or setting the same BERT encoder weights for the decoder is not optimal since it results in a model whose encoder and decoder were not pre-trained together to generate text.
Fine-tuning a multilingual seq2seq network is a better approach since the model is pre-trained for text generation as a whole.
Yet it is not optimal either, as multilingual models are exposed to a trace amount of Turkish during pre-training, and consequently, most of the training time and learning capacity of the network are used for high-resource languages.

Therefore, both of the approaches so far rely on sub-optimal pre-trained models and underachieve the fine-tuning potential.

\section{Model}
\label{sec:vbart}

\subsection{Tokenizer}
Before training a model from scratch, a tokenizer is needed first.
We use SentencePiece~\cite{sentencepieceframework_kudo2018sentencepiece} Unigram Model Tokenizer that is part of VNLP~\cite{vnlpturker2024vnlp}  library.
It was trained on 10 GB of text that consists of random subsets of OSCAR~\cite{oscar_2022arXiv220106642A}, OPUS~\cite{opus_zhang2020improving} and Wikipedia dump corpora.
Tokenizer training required 500+ GBs of memory and took 3 hours on 96 CPU cores.

\subsection{Network Architecture}
The network is based on mBART architecture rather than BART.
Following the finding of mBART authors, this is a conscious design choice and serves to stabilize the FP16 mixed precision training~\cite{fp16_micikevicius2017mixed} thanks to post-encoder and post-decoder LayerNorm layers.
The only difference from the mBART architecture is that we use sinusoidal positional embeddings.
The network follows the original BART-Large configuration with 12 encoder and decoder layers, 16 heads and 1024 model dimension.
Having a vocabulary size of 32,000 in this configuration results in 387.6M trainable parameters.

\subsection{Pre-training Task}
The pre-training task is sentence permutation combined with span masking where 30\% of tokens are masked with span length defined by the Poisson distribution ($\lambda = 3.5$), following the BART-Large objective.

\subsection{Training Corpus}
Training corpus is made of Turkish sections of OSCAR-2201 \cite{oscar_2022arXiv220106642A} and mC4 \cite{mt5mc4_xue2020mt5}, which contain 10.8M and 87.7 million pages, respectively.
They are concatenated to obtain the final training corpus of 98.5M pages in total.
However, web-crawled data is often noisy and full of keywords, titles and other non-sentence texts.
In order to obtain a higher-quality dataset, these pages are cleaned using a chain of rules and heuristics.
See the Appendix
for the details of the data cleaning process.
The cleaned corpus holds 135.7 GB of space on disk, contains 50.3M pages and is made of 25.33B subword tokens in total.
We call this curated dataset \emph{vngrs-web-corpus}.

\subsection{Data Generator}
A data generator is written to feed the model training.
It works dynamically so that even when the same text input is read from the corpus, stochastic text noising processes result in a different permutation of sentences and tokens masked.
Moreover, in the case of samples that are longer than the context length, a random continuous span of sentences that will fit is selected. This has two benefits. First, no sentence is cut in half. Second, the model gets to see other sections of longer texts as well.

Both encoder and decoder context lengths are set to 1024 with right padding.
However, noising of the encoder inputs results in shorter sequences since spans of tokens are replaced by a single mask token.
This results in a waste of computation on the encoder side.
Hence, the encoder context length is set to 800 during pre-training for efficiency.

\begin{table*}[]
\centering
\resizebox{\textwidth}{!}{

\begin{tabular}{lccccccc}
 & & \multicolumn{3}{c}{From News Summary} & \multicolumn{3}{c}{From News Content} \\
\scriptsize R1/R2/RL & Parameters & MLSum & TRNews & XLSum & MLSum & TRNews & XLSum \\
\hline
\multicolumn{8}{l}{\textbf{\citealp{baykara_seq2seq_baykara2022turkish}} } \\
\hline
BERT2BERT-32K & 248M & 39.35/21.14/37.55 & 41.87/24.37/40.88 & & & & \\
mBART25 & 610M & 34.85/18.03/33.46 & 37.72/20.99/36.74 & & &  & \\
mT5-Base     & 580M & 40.77/22.42/38.97 & 41.87/24.49/40.87 & & & & \\
\hline
\multicolumn{8}{l}{\textbf{Our work}} \\
\hline
VBART-Large  & 387M & \textbf{45.17/30.49/42.92} & \textbf{43.05/29.20/41.87} & 42.72/26.79/40.70 & 39.82/25.88/37.83 & 39.26/25.45/38.11 & 37.40/21.09/34.71 \\
VBART-XLarge  & 740M & \textbf{45.11/30.57/42.98} & \textbf{43.52/29.56/42.16} & 42.89/26.92/40.70 & 39.79/25.88/37.79 & 39.41/25.61/38.25 & 37.39/21.20/34.87 \\

\end{tabular}
}
\caption{Title Generation}
\label{tbl:tg}
\end{table*}

\subsection{Training}
Training is carried out on 8X Nvidia A100-80 GB on AWS for 2.7M steps with a batch size of 256 and took 30 days.
Adam~\cite{adamoptimizer_kingma2014adam} optimizer ($\beta_{1} = 0.9, \beta_{2} = 0.98, \epsilon = 1e-6$) along with the learning rate scheduler (20,000 warmup steps) from the original Transformer paper is used.
Dropout is set to 0.1 for the first 2.33M steps, reduced to 0.05 for the next 165K steps and finally 0 for the last 205K steps.

Considering that the model context length is 1024 tokens and training is carried out with a batch size of 256 on a dataset that is made of 25.33B tokens, 2.7M steps correspond to 28 epochs.
Corresponding training loss can be seen in Figure~\ref{fig:train_loss}.

\subsection{Model Enlargement}
After the pre-training of VBART-Large, we decided to create an XLarge version by doubling the number of encoder and decoder layers while keeping the other configuration intact.
Having a smaller budget, VBART-XLarge is initialized by using VBART-Large weights to speed up the training.
All weights are interchangeable except for the newer Encoder and Decoder layers.
Thus, we set every odd-numbered encoder and decoder layer weights from the Large model while initializing even-numbered layer weights from scratch.

Then VBART-XLarge is pre-trained on the same hardware and the pre-training task for 8 days.
Due to time constraints and the increased model size, the number of warmup steps and batch size are reduced to 5,000 and 128, respectively.
Thanks to weight transfer, the XLarge model reached the loss values that the Large model had reached in 1 and 20 days in 1.5 hours and 3 days, respectively.
In total, the XLarge model is trained for 640K steps, with the first 480K steps with a Dropout value of 0.1, the next 80K steps with a Dropout value of 0.05, and the final 80K steps with a Dropout value of 0.

\subsection{Implementation}
SentencePiece~\cite{sentencepieceframework_kudo2018sentencepiece} library is used to train the Unigram Model tokenizer.
The network, data generator and training are implemented using Tensorflow~\cite{tensorflow_tensorflow2015} framework.
Moreover, trained networks and the tokenizer are converted to corresponding Huggingface implementations in order to use Huggingface Transformers'~\cite{huggingface_transformers_wolf2019huggingface}text generation utilities.
They can be accessed on the Huggingface hub
\footnote{\href{https://huggingface.co/vngrs-ai}{https://huggingface.co/vngrs-ai}}
along with the vngrs-web-corpus.

\begin{table*}[]
\centering
\resizebox{\textwidth}{!}{
\centering
\begin{tabular}{lccc}
 &Parameters&Open Subtitles&Tatoeba\\
\hline \multicolumn{2}{l}{\textbf{\citealp{alkurdi-etal-2022-semantic}} }& \multicolumn{2}{c}{BERTScore-\{cased/uncased\} /  BLEU / Rouge-L / METEOR / TER} \\ \hline
mT5-Base Open Subtitles& \multirow{2}*{580M}    & 88.89/91.94/36.40/73.87/72.16/37.58& 91.61/93.93/34.74/86.60/84.85/18.23\\ 
mT5-Base Tatoeba& &88.95/92.08/38.13/68.39/65.87/45.13& 91.97/94.20/37.02/84.05/81.59/22.76\\ 
\hline \multicolumn{4}{l}{\textbf{Our work}}\\ \hline
 VBART-Large&387M& \textbf{89.25}/\textbf{92.22}/\textbf{47.78}/\textbf{75.35}/\textbf{73.93}/\textbf{35.74} & \textbf{95.79}/\textbf{96.87}/\textbf{69.86}/\textbf{88.14}/\textbf{86.56}/\textbf{16.93} \\
 VBART-XLarge&740M&\textbf{89.30}/\textbf{92.25}/\textbf{47.29}/\textbf{75.38}/\textbf{74.04}/\textbf{35.72} & \textbf{95.66}/\textbf{96.77}/\textbf{68.21}/\textbf{87.56}/\textbf{86.14}/\textbf{17.51}\\
\end{tabular}}
\caption{Text Paraphrasing}
\label{tbl:pp}
\end{table*}

\section{Experiments}
\label{sec:experiments}
After pre-training, VBART models are fine-tuned on various tasks using Adam optimizer ($\beta_{1} = 0.9, \beta_{2} = 0.98, \epsilon = 1e-6$) with different learning rates varying by the downstream task, applying a linear decay of 0.95.
Each task's evaluation is conducted using the referenced work's evaluation script for consistency.

\subsection{Text Summarization}
Large and XLarge models are fine-tuned on Turkish sections of MLSum~\cite{mlsum_scialom2020mlsum}, TRNews~\cite{baykara_seq2seq_baykara2022turkish}, XLSum~\cite{xlsum_hasan-etal-2021-xl} and Wikilingua~\cite{wikilingua_ladhak2020wikilingua} datasets for 30 and 20 epochs respectively, with a learning rate of $1e-5$.
Then they are evaluated on the corresponding test splits.
Note that Wikilingua does not contain a test split for Turkish, so it is excluded from the evaluation.

Rouge-1, Rouge-2 and Rouge-L scores are computed, and a comparison with the previous work is reported in Table~\ref{tbl:sum}.
It is the higher the better for Rouge metrics.
It is observed that our models surpass the previous state-of-the-art in these datasets on 5 out of 6 metrics, despite the Large model having significantly fewer parameters.
The only exception to this is BERT2BERT-32K model, which achieved better Rouge-L on the TRNews dataset.
Moreover, we report the results on the XLSum dataset for the first time in the literature in this work.

\subsection{Title Generation}
VBART models are fine-tuned to generate titles from given news and summaries.
The cartesian product of model sizes and training data results in 4 distinct models.
Large model is fine-tuned for 15 and 25 epochs on news summary and content as input, while XLarge model is fine-tuned for 10 and 15 epochs.

Similar to text summarization, Rouge-1, Rouge-2 and Rouge-L are used for evaluation.
Comparison with the earlier work is reported in Table~\ref{tbl:tg}.
Our models surpass the previous work by far in all metrics and datasets.
Since \citet{baykara_seq2seq_baykara2022turkish} did not fine-tune any models to generate titles from the news content, there is no result to compare.

\begin{figure*}[]
\centering
\includegraphics[width=1.3\columnwidth]{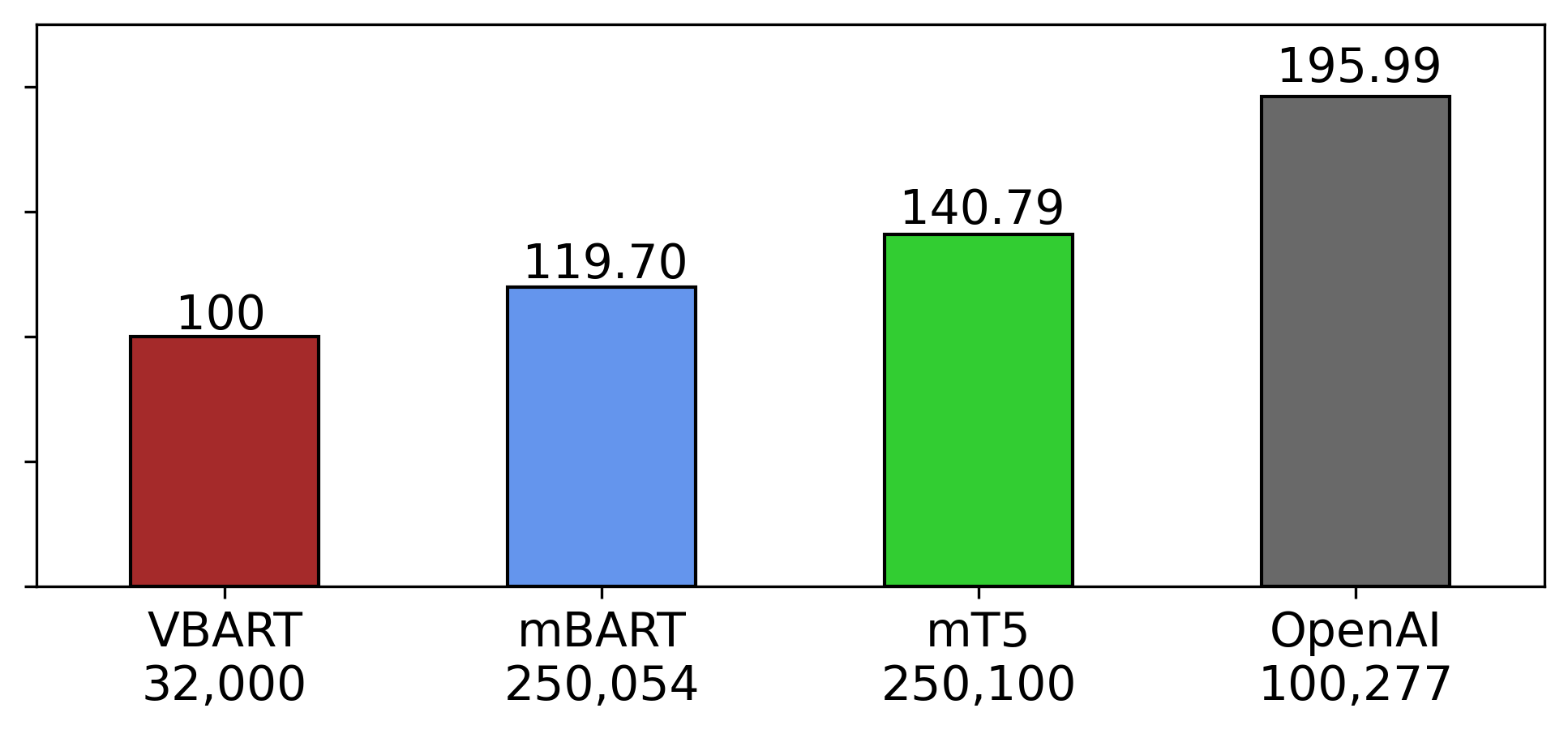}
\caption{Relative number of tokens spent to encode Turkish text in average.
The numbers below show the vocabulary size of each tokenizer. Despite their large vocabulary size, multilingual tokenizers spend up to 96\% more tokens than VBART Tokenizer.
}
\label{fig:relative_num_tokens}
\end{figure*}

\subsection{Text Paraphrasing}
VBART models are fine-tuned for 20 and 25 epochs, respectively, on a mixture of OpenSubtitles~\cite{opensubtitles_lison2016opensubtitles2016}, TED~\cite{ted_cettolo2012wit3} and Tatoeba~\cite{tatoeba_artetxe2019massively} datasets mentioned in the paper published by \citet{alkurdi-etal-2022-semantic}.
The original paper suggests a method to improve the dataset quality. Since the filtered data is not published, our models are trained and evaluated on the unfiltered data.
Then, we compare our models to the models that were fine-tuned on unfiltered data.
Results are evaluated using BERTScore, BLEU, Rouge-L, Meteor and Translation Error Rate (TER) metrics. BERTScore is computed using cased
\footnote{\href{https://huggingface.co/dbmdz/bert-base-turkish-cased}{dbmdz/bert-base-turkish-cased} last reached in \textit{19 Oct 2023}} 
and uncased
\footnote{\href{https://huggingface.co/dbmdz/bert-base-turkish-uncased}{dbmdz/bert-base-turkish-uncased} last reached in \textit{19 Oct 2023}} 
versions.
Results are reported in Table~\ref{tbl:pp}.
Except for TER, which is the lower, the better; for all metrics, the higher, the better.

We compare our models with the two models from the referenced work.
Our models outperform both mT5-Base models in every metric, especially by a large margin for BERTScore.

\begin{table*}[]
\centering
\resizebox{\textwidth}{!}{

\begin{tabular}{lccccccc}
          &            & \multicolumn{3}{c}{Question Generation Task}               & \multicolumn{3}{c}{Question Answering Task}              \\
          &            & \multicolumn{3}{c}{BLEU-1 / BLEU-2 / Rouge-L}             & \multicolumn{3}{c}{F1 / EM}                               \\
          & Parameters & TQuADv1           & TQuADv2           & XQuAD             & TQuADv1           & TQuADv2           & XQuAD             \\ \hline
\multicolumn{8}{l}{\textbf{\citealp{AKYN2022}} }                                                                                                   \\ \hline
mT5-Small    & 300M & 37.3/30.1/44.3  & 39.6/32.9/46.5 & 21.1/13.8/28.4 & 63.8/48.5  & 67.1/50.5 & 48.8/32.9 \\
mT5-Base     & 580M & 48.4/41.7/53.6  & 47.6/41.2/53.9 & 27.9/20.9/35.8 & 72.1/55.8 & 71.5/56.2 & 61.1/43.3 \\
mT5-Large    & 1.2B & \textbf{49.8}/\textbf{43.2}/\textbf{55.2} & 49.1/42.7/54.3 & 29.3/21.9/\textbf{37.5}& 74.7/\textbf{59.6} & 73.3/58.4 & \textbf{65.0}/\textbf{46.7} \\ \hline
\multicolumn{8}{l}{\textbf{Our work}}                                                                                                                \\ \hline
VBART-Large  & 387M & 49.2/42.4/54.8 &
\textbf{50.8}/\textbf{43.8}/\textbf{56.5} &
\textbf{30.0}/\textbf{22.8}/35.2 & 73.9/58.5& \textbf{75.6}/\textbf{59.6} & 60.2/44.0 \\
VBART-XLarge  & 740M & 49.1/42.2/54.9 &
\textbf{51.5}/\textbf{44.7}/\textbf{57.1} &
\textbf{30.5}/\textbf{23.1}/35.7 & \textbf{75.1}/59.4& \textbf{77.1}/\textbf{61.0} & 62.4/\textbf{46.7} \\

\end{tabular}
}
\caption{Question Generation and Answering}
\label{tbl:qa}
\end{table*}

\subsection{Question Generation and Answering \label{sect:qa-qg}}
Our models are fine-tuned on "Turkish NLP Q{\&}A Dataset" \cite{tquad} (will be abbreviated as TQuAD) for three tasks - answer extraction, question generation and question answering- using the methodology described in \citet{AKYN2022}.
TQuAD has two variations, TQuADv1 and TQuADv2, with a large intersection of samples.
Moreover, there are some samples that exist in both the train and the evaluation sets. These intersections are reported in the Appendix~\ref{sec:qa-qg-dataset}
We convert the prompts to Turkish and process the raw data using the script from the referenced work's github repository \footnote{ \href{https://github.com/obss/turkish-question-generation}{OBSS/turkish-question-generation} last reached in \textit{19 Sep 2023}.}.

We fine-tune our Large and XLarge models for 50 and 55 epochs, respectively, with a learning rate of $5e-6$. 
The train set is obtained by concatenating TQuaDv1 and TQuaDv2 and dropping the duplicate samples.

Then we evaluate on the test splits of TQuADv1, TQuADv2 and XQuAD~\cite{xquad} and report the results on Table\ref{tbl:qa}.
BLEU-1, BLEU2 and Rouge-L are computed for question generation task.
F1 and Exact Match (EM) are computed for question answering task.
In order to be comparable, we do not drop the intersected test samples from the training dataset, following the original work.
Also, note that following \citet{AKYN2022}, we report metrics up to a single decimal in order to be comparable.
Therefore, these results are useful for comparison only and are not realistic enough for the tasks themselves.

Our models outperform mT5-Base and mT5-Small and are comparable to mT5-Large, which is a 1.2B parameters model.
Out of 15 metrics, VBART-Large and VBART-XLarge models surpass mT5-Large in 7 and 8 metrics, respectively.
Overall, mT5-Large is better on TQuADv1, while ours are better on TquADv2.

\section{Discussion}
\subsection{Tokenizer}
Although one often thinks of the number of model parameters when comparing computational efficiency, it is equally crucial to be able to represent text using fewer tokens.
Tokenizers of multilingual models are trained on multilingual corpora.
This has two consequences.
First, their vocabulary size is chosen to be large to be comprehensive enough.
Second, despite their large vocabulary size, it takes more tokens to encode a text in a language compared to a tokenizer that was trained in the specific language alone.

We compare VBART Tokenizer to mBART, mT5 and OpenAI's cl100k\_base tokenizers.
In doing so, we randomly sample 1M pages from vngrs-web-corpus and tokenize each page.
Then we count the total number of tokens generated by each tokenizer.
Figure~\ref{fig:relative_num_tokens} shows the relative number of tokens required to encode Turkish text, normalized by VBART Tokenizer.
Despite having almost 8x vocabulary size, mBART and mT5 tokenizers spend 19.70\% and 40.79\% more tokens, respectively.
OpenAI's cl100k\_base tokenizer, which is used along with Chat-GPT and GPT-4 models, spends 95.99\% more tokens despite its more than 3x vocabulary size.

Moreover, taking vocabulary sizes into account, we can claim that VBART Tokenizer's representation power is $\frac{119.70}{100} \times \frac{250,054}{32,000} = 9.35$ times mBART's.
This is 11.00 and 6.14 for mT5 and OpenAI tokenizers, respectively.


\subsection{Experiments}
Fine-tuning results of 4 downstream tasks show that having a dedicated pre-trained model for a language outperforms multilingual mT5 and mBART models, even when the number of parameters is significantly less for the prior.
VBART-Large model is able to surpass mBART25, mBART50 and mT5-Base models despite having 37\% and 33\% less parameters, respectively.
Moreover, it is observed in Table~\ref{tbl:qa} that VBART-Large is at par with mT5-Large, which has more than 3x parameters.

\subsection{VBART-Large vs. VBART-XLarge}
It is evident that the XLarge model's improvement over the Large model is small.
This is because the XLarge model is pre-trained for far fewer steps than the Large one.
The Large model is exposed to 708B tokens, while this is only 84B for the XLarge version, 88\% less.
Despite that, the XLarge version is able to improve the results of the Large, although with small margins.
We hypothesize that the XLarge version can improve significantly when pre-trained for more steps.

\subsection{Chinchilla Scaling Law}
Chinchilla scaling law~\cite{chincilla_hoffmann2022training} states that, in the case of single epoch training, an LLM is optimal if there are 20 tokens in the training set per network parameter.
Then VBART-Large and VBART-XLarge are
$\frac{708B \text{ tokens}}{387.6M \times 20} = 91.33$
and
$\frac{84B \text{ tokens} \times 28 \text{ epochs}}{387.6M \times 20} = 10.83$
fold optimal, respectively.
Accordingly, Large and XLarge models should have been pre-trained for 30K and 59K steps, respectively.
However, we observe significant improvements over the course of pre-training in Figure~\ref{fig:train_loss}.

Moreover, unlike auto-regressive GPT~\cite{gpt1_radford2018improving}
models, pre-training objective and the dynamic data generator used in this work provide different input-output pairs every time a page is sampled, augmenting the dataset.

Considering these, it is a matter of question whether the Chinchilla Scaling Law is applicable to encoder-decoder models or models with dynamic data generators with a pre-training objective other than next token prediction.

\subsection{Future Work}
This work can be extended in multiple ways for the monolingual setting.
The most trivial direction is further model enlargement by increasing 24 layers of the XLarge model to 36 layers and hitting 1B parameters.
Others are pre-training another Turkish model based on a different architecture, using a different pre-training objective such as T5, or on a larger dataset.

Beyond language-specific directions, there are two research directions that can be explored.
First, the model enlargement technique proposed in this work can be exclusively studied. This is a promising direction that can considerable amount of training costs.
Second, the applicability of Chinchilla Scaling Law to any other than decoder-only models, the pre-training task of which is next-token prediction, is an open question. Despite the recent popularity of decoder-only models, it would be beneficial to have a heuristic to determine the pre-training configuration of an encoder-decoder model.

\section{Conclusion}
\label{sec:conclusion}
In this work, we presented the first sequence-to-sequence LLMs for the Turkish language.
All of the components; dataset, tokenizer and models are prepared and trained for Turkish from scratch.
We showed our tokenizer's ability of compact representation over multilingual ones, compared the two models proposed in this work, introduced a method to enlarge an existing LLM and questioned the relevancy of Chinchilla Scaling Law to encoder-decoder models.
Achieving new state-of-the-art results for abstractive text summarization, paraphrasing, title generation, question answering and question generation, we showed that monolingual pre-trained models surpass multilingual ones when fine-tuned for downstream tasks, despite having significantly fewer parameters.

\section*{Acknowledgements}
We thank Amazon Web Services for funding the pre-training of the models.

\cleardoublepage

\section{Bibliographical References}\label{sec:reference}
\bibliography{bibliography}
\bibliographystyle{lrec-coling2024-natbib}

\cleardoublepage
\appendix

\section{Data Cleaning}
\label{sec:appendix_data_cleaning}
Data cleaning is conducted on two levels: page cleaning and sentence cleaning.
T5~\cite{t5_2019t5} authors describe their data cleaning process in detail.
In this work, they are leveraged and extended for a cleaner dataset.

\subsection{Page Cleaning}
\subsubsection{Anomaly Detection}
Some of the pages are filled with Search Engine Optimization (SEO) targeted, repeating keywords, titles and other entities that do not make a sequence of sentences that belong to a context.
In order to get rid of these pages in an unsupervised way, Isolation Forest anomaly detection algorithm~\cite{isolationforest_liu2008isolation}
is used on five heuristics that separate the mentioned bad pages from the good ones.
They are:
\begin{itemize}
    \item Mean sentence length on the page
    \item Standard deviation of sentence length on the page
    \item Maximum sentence length on the page
    \item Ratio of short sentences(less than 4 words) to the number of sentences in the page
    \item Ratio of uppercase characters to the number of characters on the page
\end{itemize}
Then, the model is trained and inferred on the whole dataset with default configuration to obtain the anomaly scores.

\subsubsection{Rule-based Cleaning}
After obtaining anomaly scores, final page filtering is applied by removing pages
\begin{itemize}
    \item whose anomaly score is below 0.05
    \item that contains bad/naughty words
    \item that contains "lorem ipsum"
    \item whose language probability is lower than 0.85 (applicable for OSCAR corpus only)

\end{itemize}

This results in a 19.81\% reduction in the number of pages.

\subsection{Sentence Cleaning}
Among the remaining pages, each page is split into sentences, and each sentence is discarded if any of the following is True:

\begin{itemize}
  \item It is an empty string
  \item It does not end with sentence termination punctuation
  \item It contains curly bracket "\{" or "\}"
  \item It contains "JavaScript"
  \item It contains "gizlilik ve çerezler" (privacy and cookies)
  \item It contains "|"
  \item The number of words in the sentence is less than 4 or more than 50
  \item The longest word in the sentence is longer than 30 characters, which is the 0.995 quantile
  \item Capital letters account for more than half of the sentence
  \item 1/3 of the characters are numeric
  \item It does not contain any punctuation
  \item It contains too many duplicate words; that is, the average number of duplicate words is greater than 2
\end{itemize}

\subsection{Finalization of the Dataset}
As the last step, pages with less than 5 sentences are cleaned, which further reduces the number of pages by 28.69M.
In the end, 50.3M pages consisting of 25.33B subword tokens remain to form the 135.7GB dataset.

\section{Question Generation \& Answering Dataset Intersections}
\begin{table}[h]
\label{tab:intersections}
\begin{subtable}[h]{\columnwidth}
\centering
\caption{Raw (JSON)}
\label{tab:json-intersection}
\resizebox{.95\columnwidth}{!}{%
\begin{tabular}{c|c|ccc}
\cline{2-5}
 &
  \begin{tabular}[c]{@{}c@{}}xquad-eval\\ (1190)\end{tabular} &
  \multicolumn{1}{c|}{\begin{tabular}[c]{@{}c@{}}tquad2-eval\\ (3034)\end{tabular}} &
  \multicolumn{1}{c|}{\begin{tabular}[c]{@{}c@{}}tquad2-train\\ (14221)\end{tabular}} &
  \multicolumn{1}{c|}{\begin{tabular}[c]{@{}c@{}}tquad1-eval\\ (2612)\end{tabular}} \\ \hline
\multicolumn{1}{|c|}{\begin{tabular}[c]{@{}c@{}}tquad1-train\\ (8308)\end{tabular}} &
  0 &
  \multicolumn{1}{c|}{194} &
  \multicolumn{1}{c|}{7641} &
  \multicolumn{1}{c|}{194} \\ \hline
\multicolumn{1}{|c|}{\begin{tabular}[c]{@{}c@{}}tquad1-eval\\ (2612)\end{tabular}} &
  0 &
  \multicolumn{1}{c|}{882} &
  \multicolumn{1}{c|}{194} &
   \\ \cline{1-4}
\multicolumn{1}{|c|}{\begin{tabular}[c]{@{}c@{}}tquad2-train\\ (14221)\end{tabular}} &
  86 &
  \multicolumn{1}{c|}{194} &
   &
   \\ \cline{1-3}
\multicolumn{1}{|c|}{\begin{tabular}[c]{@{}c@{}}tquad2-eval\\ (3034)\end{tabular}} &
  0 &
   &
   &
   \\ \cline{1-2}
\end{tabular}%
}
\end{subtable}

\vspace{1em} 

\begin{subtable}[h]{\columnwidth}
\centering
\caption{Processed (Script output)}
\label{tab:dataset-intersection}
\resizebox{.95\columnwidth}{!}{%
\begin{tabular}{c|c|ccc}
\cline{2-5}
 &
  \begin{tabular}[c]{@{}c@{}}xquad-eval\\ (5184)\end{tabular} &
  \multicolumn{1}{c|}{\begin{tabular}[c]{@{}c@{}}tquad2-eval\\ (6368)\end{tabular}} &
  \multicolumn{1}{c|}{\begin{tabular}[c]{@{}c@{}}tquad2-train\\ (75380)\end{tabular}} &
  \multicolumn{1}{c|}{\begin{tabular}[c]{@{}c@{}}tquad1-eval\\ (3765)\end{tabular}} \\ \hline
\multicolumn{1}{|c|}{\begin{tabular}[c]{@{}c@{}}tquad1-train\\ (45069)\end{tabular}} &
  0 &
  \multicolumn{1}{c|}{520} &
  \multicolumn{1}{c|}{20329} &
  \multicolumn{1}{c|}{520} \\ \hline
\multicolumn{1}{|c|}{\begin{tabular}[c]{@{}c@{}}tquad1-eval\\ (3765)\end{tabular}} &
  0 &
  \multicolumn{1}{c|}{2343} &
  \multicolumn{1}{c|}{520} &
   \\ \cline{1-4}
\multicolumn{1}{|c|}{\begin{tabular}[c]{@{}c@{}}tquad2-train\\ (75380)\end{tabular}} &
  227 &
  \multicolumn{1}{c|}{520} &
   &
   \\ \cline{1-3}
\multicolumn{1}{|c|}{\begin{tabular}[c]{@{}c@{}}tquad2-eval\\ (6368)\end{tabular}} &
  0 &
   &
   &
   \\ \cline{1-2}
\end{tabular}

}
\end{subtable}
\caption{Data intersections}

\end{table}

\label{sec:qa-qg-dataset}
Working on question-answering and question-generation tasks, we have observed that there are intersected samples in the splits that exist in other splits as well, including between train and evaluation splits.

In order to understand the extent and cause of the intersections, we have examined raw and processed versions, conducting a cross-examination.
We report the results in Table~\ref{tab:json-intersection} and Table~\ref{tab:dataset-intersection}.
Cell values are the number of samples intersected between the corresponding datasets.
The numbers under the dataset names indicate the total number of samples in the dataset.

\end{document}